\documentclass{article}

    \usepackage[preprint,nonatbib]{neurips_2022}

\usepackage{xr-hyper}

\usepackage[utf8]{inputenc} % allow utf-8 input
\usepackage[T1]{fontenc}    % use 8-bit T1 fonts
\usepackage{hyperref}       % hyperlinks
\usepackage{url}            % simple URL typesetting
\usepackage{booktabs}       % professional-quality tables
\usepackage{amsfonts}       % blackboard math symbols
\usepackage{nicefrac}       % compact symbols for 1/2, etc.
\usepackage{microtype}      % microtypography
\usepackage{xcolor}         % colors

\usepackage{main}
\usepackage[citestyle=authoryear,sorting=nyt,maxnames=1,uniquelist=false]{biblatex}
\usepackage{graphicx}
\graphicspath{ {./figures/} }

\usepackage{algorithm}      % http://ctan.org/pkg/algorithms
\usepackage{algpseudocode}  % http://ctan.org/pkg/algorithmicx

\usepackage{amsmath}

\DeclareMathOperator*{\LLM}{LLM}

\usepackage{wrapfig}
\usepackage{subcaption}
\usepackage{array}          % tt inside tabular
\usepackage{listings}
\usepackage{arydshln}       % dash lines in table

\NewDocumentCommand{\Algorithm}{}{ICPI}
\newcommand\NoArgMax{``No $\argmax$''}
\newcommand\NoBalance{``No Balance''}
\newcommand\NoConstraints{``No Constraints''}
\newcommand\CSixteen{``$c=16$''}
\newcommand\NoHints{``No Hints''}
\newcommand{\MatchingModel}[1][]{``Matching Model''}
\newcommand{\TabularQ}[1][]{``Tabular Q''}
\title{Large Language Models \\ Can Implement Policy Iteration}

\author{%
Ethan Brooks\textsuperscript{\rm 1},
Logan Walls\textsuperscript{\rm 2}, 
Richard L. Lewis\textsuperscript{\rm 2}, 
Satinder Singh\textsuperscript{\rm 1} \\
\textsuperscript{\rm 1}Computer Science and Engineering, University
of Michigan \\
\textsuperscript{\rm 2}Department of Psychology, University of Michigan \\
\texttt{\{ethanbro,logwalls,rickl,baveja\}@umich.edu} \\
}

\begin{document}

\maketitle

\begin{abstract}
  In this work, we demonstrate a method for implementing policy iteration using
  a large language model. While the application of foundation models to RL has
  received considerable attention, most approaches rely on either (1) the
  curation of expert demonstrations (either through manual design or
  task-specific pretraining) or (2) adaptation to the task of interest using
  gradient methods (either fine-tuning or training of adapter layers). Both of
  these techniques have drawbacks. Collecting demonstrations is labor-intensive,
  and algorithms that rely on them do not outperform the experts from which the
  demonstrations were derived. All gradient techniques are inherently slow,
  sacrificing the “few-shot” quality that makes in-context learning attractive
  to begin with. Our method demonstrates that a large language model can be used
  to implement policy iteration using the machinery of in-context learning,
  enabling it to learn to perform RL tasks without expert demonstrations or
  gradients. Our approach iteratively updates the contents of the prompt from
  which it derives its policy through trial-and-error interaction with an RL
  environment. In order to eliminate the role of in-weights learning (on which
  approaches like Decision Transformer rely heavily), we demonstrate our
  method using Codex \parencite{chen2021evaluating}, a language model with no
  prior knowledge of the domains on which we evaluate it.
\end{abstract}

\section{Introduction}
% Large Language Models (LLMs) demonstrate powerful sequence completion abilities,
% detecting patterns in the input prompt, inferring their underlying logic, and
% generalizing this logic to produce sensible continuations. In addition these
% models distill a tremendous body of knowledge relating to the empirical world.
% In this work, we demonstrate that these properties can be leveraged to implement
% policy iteration, enabling the model to learn to perform RL tasks entirely in
% context without any updates to the model weights. While the tasks demonstrated
% here are simple, we expect that approaches like this will scale as

In many settings, models implemented using a transformer or recurrent
architecture will improve their performance as information accumulated in their
context or memory. We refer to this phenomenon as ``in-context learning.''
\parencite{brown2020language} demonstrated a technique for inducing this phenomenon
by prompting a large language model with a small number of input/output
exemplars. An interesting property of in-context learning in the case of large
pre-trained models (or ``foundation models'') is that the models are not
directly trained to optimize a meta-learning objective, but demonstrate an
emergent capacity to generalize (or at least specialize) to diverse downstream
task-distributions \parencite{wei_emergent_2022}.

A litany of existing work has explored methods for applying this remarkable
capability to downstream tasks (see \nameref{sec:related}), including
Reinforcement Learning (RL). Most work in this area either (1) assumes access to
expert demonstrations --- collected either from human experts
\parencite{huang_language_2022,baker_video_2022}, or domain-specific pre-trained RL
agents
\parencite{chen_decision_2021,lee_multi-game_2022,janner_offline_2021,reed_generalist_2022,xu_prompting_2022}.
--- or (2) relies on gradient-based methods --- e.g. fine-tuning of the
foundation models parameters as a whole
\parencite{lee_multi-game_2022,reed_generalist_2022,baker_video_2022} or newly
training an adapter layer or prefix vectors while keeping the original
foundation models frozen
\parencite{li_prefix-tuning_2021,singh_know_2022,karimi_mahabadi_prompt-free_2022}.

Our work demonstrates an approach to in-context learning which relaxes these
assumptions. Our method, In-Context Policy Iteration (\Algorithm{}), implements
policy iteration using the prompt content, instead of the model parameters, as
the locus of learning, thereby avoiding gradient methods. Furthermore, the use
of policy iteration frees us from expert demonstrations because suboptimal
prompts can be improved over the course of training.

We illustrate the method empirically on six small illustrative RL tasks---
\emph{chain, distractor-chain, maze, mini-catch, mini-invaders}, and
\emph{point-mass}---in which the method very quickly finds good policies.  We
also compare five pretrained Large Language Models (LLMs), including two
different size models trained on natural language---OPT-30B and GPT-J---and
three different sizes of a model trained on program code---two sizes of Codex as
well as InCoder. On our six domains, we find that only the largest model (the
\texttt{code-davinci-001} variant of Codex) consistently demonstrates learning.

\begin{figure*}[t]
  \centering%
  \includegraphics[width=\textwidth]{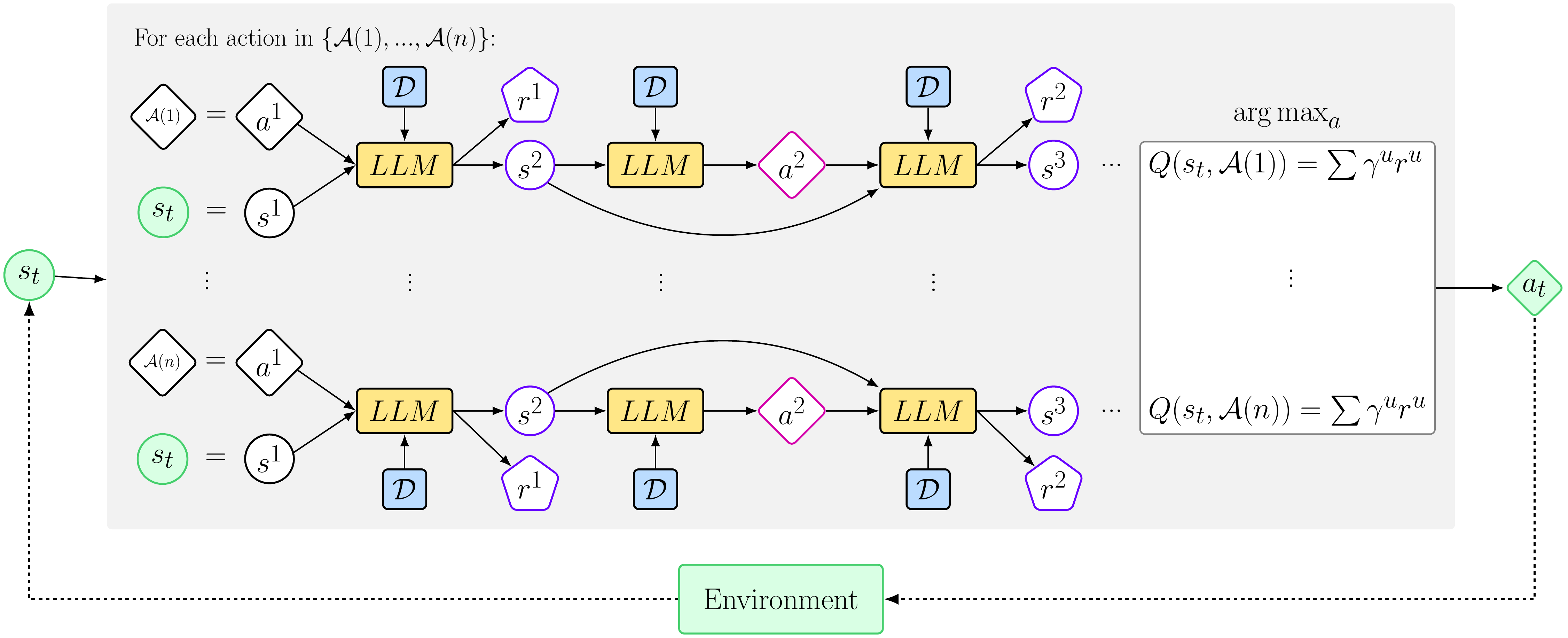}
  \caption{For each possible action $\Actions(1), \dots, \Actions(n)$, the LLM
    generates a rollout by alternately predicting transitions
    and selecting actions. Q-value estimates are discounted
    sums of rewards. The action is chosen greedily with respect to Q-values.
    Both state/reward prediction and next action selection use trajectories from $\Buffer$ to create prompts for the LLM. Changes to the content of $\Buffer$ change the prompts that the LLM receives, allowing the model to improve its behavior over time.}
  \label{fig:q rollouts}
\end{figure*}

\section{Related Work}
\label{sec:related}

A common application of foundation models to RL involves tasks that have
language input, for example natural language instructions/goals
\parencite{garg_lisa_2022,hill_human_2020}  or text-based games
\parencite{peng_inherently_2021,singh_pre-trained_2021,majumdar_improving_2020,ammanabrolu_learning_2021}.
Another approach encodes RL trajectories into token sequences, and processes
them with a foundation model, model representations as input to deep RL
architectures
\parencite{li_pre-trained_2022,tarasov_prompts_2022,tam_semantic_2022}. Finally, a
recent set of approaches treat RL as a sequence modeling problem and use the
foundation models itself to predict states or actions. This section will focus
on this last category.

\subsection{Learning from demonstrations}
Many recent sequence-based approaches to reinforcement learning use
demonstrations that come either from human experts or pretrained RL agents. For
example, \cite{huang_language_2022} use a frozen LLM as a planner for everyday
household tasks by constructing a prefix from human-generated task instructions,
and then using the LLM to generate instructions for new tasks. This work
is extended by \cite{huang_inner_2022}. Similarly,
\cite{ahn_as_2022} use a value function that is trained on human
demonstrations to rank candidate actions produced by an LLM.
\cite{baker_video_2022} use human demonstrations to train the foundation model
itself: they use video recordings of human Minecraft players to train a
foundation models that plays Minecraft. Works that rely on pretrained RL agents
include \cite{janner_offline_2021} who train a ``Trajectory Transformer'' to
predict trajectory sequences in continuous control tasks by using trajectories
generated by pretrained agents, and \cite{chen_decision_2021}, who use a
dataset of offline trajectories to train a ``Decision Transformer'' that
predicts actions from state-action-reward sequences in RL environments like
Atari. Two approaches build on this method to improve generalization:
\cite{lee_multi-game_2022} use trajectories generated by a DQN agent to train a
single Decision Transformer that can play many Atari games, and
\cite{xu_prompting_2022} use a combination of human and artificial trajectories
to train a Decision Transformer that achieves few-shot generalization on
continuous control tasks. \cite{reed_generalist_2022} take task-generality a
step farther and use datasets generated by pretrained agents to train a
multi-modal agent that performs a wide array of RL (e.g. Atari, continuous
control) and non-RL (e.g. image captioning, chat) tasks.

Some of the above works include non-expert demonstrations as well.
\cite{chen_decision_2021} include experiments with trajectories generated by
random (as opposed to expert) policies. \cite{lee_multi-game_2022} and
\cite{xu_prompting_2022} also use datasets that include trajectories generated
by partially trained agents in addition to fully trained agents. Like these
works, our proposed method (ICPI) does not rely on expert demonstrations---but
we note two key differences between our approach and existing approaches.
Firstly, ICPI only consumes self-generated trajectories, so it does not require
any demonstrations (like \cite{chen_decision_2021} with random trajectories,
but unlike \cite{lee_multi-game_2022}, \cite{xu_prompting_2022}, and the other
approaches reviewed above). Secondly, ICPI relies primarily on in-context
learning rather than in-weights learning to achieve generalization (like
\cite{xu_prompting_2022}, but unlike \cite{chen_decision_2021} \&
\cite{lee_multi-game_2022}). For discussion about in-weights vs. in-context
learning see \cite{chan_data_2022}.
% see the next  section.

\subsection{Gradient-based Training \& Finetuning on RL Tasks}
Most approaches involve training or fine-tuning foundation models on RL tasks.
For example,
\cite{janner_offline_2021,chen_decision_2021,lee_multi-game_2022,xu_prompting_2022,baker_video_2022,reed_generalist_2022}
all use models that are trained from scratch on tasks of interest, and
\cite{singh_know_2022,ahn_as_2022,huang_inner_2022} combine frozen foundation
models with trainable components or adapters. In contrast,
\cite{huang_language_2022} use frozen foundation models for planning, without
training or fine-tuning on RL tasks. Like \cite{huang_language_2022}, ICPI does
not update the parameters of the foundation model, but relies on the frozen model's
in-context learning abilities. However, ICPI gradually builds and improves the
prompts within the space defined by the given fixed text-format for
observations, actions, and rewards (in contrast to \cite{huang_language_2022},
which uses the frozen model to select good prompts from a given fixed library of
goal/plan descriptions).

\subsection{In-Context Learning}
Several recent papers have specifically studied in-context learning.
\cite{laskin2022context} demonstrates an approach to performing in-context
reinforcement learning by training a model on complete RL learning histories,
demonstrating that the model actually distills the improvement operator of the
source algorithm.
% \cite{min_rethinking_2022} demonstrates that LLMs can learn in-context, even
% when the labels in the prompt are randomized, problemetizing the conventional
% understanding of in-context learning and showing that label distribution is more
% important than label correctness. 
\cite{chan_data_2022} and
\cite{garg_what_2022} provide analyses of the properties that drive in-context
learning, the first in the context of image classification, the second in the
context of regression onto a continuous function. These papers identify various
properties, including ``burstiness,'' model-size, and model-architecture, that
in-context learning depends on. \cite{chen_relation_2022} studies the
sensitivity of in-context learning to small perturbations of the context. They
propose a novel method that uses sensitivity as a proxy for model certainty.

\begin{algorithm}[b]
  \caption{Training Loop}
  \label{algo:train}
  \begin{algorithmic}[1]
    \Function{Train}{environment}
    \State initialize $\Buffer$
    \Comment{replay buffer containing full history of behavior}
    \While{training}
    \State $\Obs_0 \gets$ Reset environment.
    \While{episode is not done}
    \State $\Act_t \gets \argmax_{a}$ \Call{$\Q$}{$\Obs_t, a, \Buffer$}
    \Comment{policy improvement}
    \State $\Obs_{t+1}, \Rew_t, \Ter_t \gets$ Execute $\Act_t$ in environment.
    \State $t \gets t + 1$
    \EndWhile
    \State $\Buffer \gets \Buffer \cup
      \left(
      \Obs_0, \Act_0, \Rew_0, \Ter_0, \Obs_1, \dots,
      \Obs_t, \Act_t, \Rew_t, \Ter_t, \Obs_{t+1}
      \right)$
    \Comment{add trajectory to buffer}
    \EndWhile
    \EndFunction
  \end{algorithmic}
\end{algorithm}

\begin{figure}[tb]
  %  \vspace*{-2.5em}
  %  \begin{minipage}{0.6\textwidth}
  \begin{algorithm}[H]
    \caption{Computing Q-values}
    \label{algo:q}
    \begin{algorithmic}[1]
      \Function{$\Q$}{$\Obs_t, \Act, \Buffer$}
      \State $u \gets t$
      \State $\Obs^1 = \Obs_t$
      \State $\Act^1 = \Act$
      \Repeat
      \Comment{All samples come from the experience buffer $\Buffer$}
      \State $\Buffer_{\Ter} \sim$ time-steps with action $\Act^u$ \Comment{balancing terminal and non-terminal}
      \State $\Ter^u \sim \LLM\left(\Buffer_{\Ter}, \Obs^u, \Act^u\right)$
      \State $\Buffer_{\Rew} \sim$ time-steps with action $\Act^u$ and termination $\Ter^u$ \Comment{balancing reward}
      \State $\Rew^u \sim \LLM \left(\Buffer_{\Rew}, \Obs^u, \Act^u\right)$
      \State $\Buffer_{\Obs} \sim$ time-steps with action $\Act^u$ and termination $\Ter^u$ \Comment{no balancing}
      \State $\Obs^{u+1} \sim \LLM \left(\Buffer_{\Obs}, \Obs^u, \Act^u\right)$
      \State $\BestTrajectories \sim \RecencyCutoff$ recent trajectories
      \State $\Act^{u+1} \sim \LLM\left(\Obs^{u+1}, \BestTrajectories\right)$
      \State $u \gets u + 1$
      \Until{$\Ter^u$ is terminal }
      \State \textbf{return} $\sum_{k=1}^{u} \gamma^{k - 1} \Rew^k$
      \EndFunction
    \end{algorithmic}
  \end{algorithm}
  \vspace{-20pt}
\end{figure}

\section{Method}
\label{sec:algorithm}

How can standard policy iteration make use of in-context learning? Policy
iteration is either \textit{model-based}---using a world-model to plan future
trajectories in the environment---or \textit{model-free}---inferring
value-estimates without explicit planning. Both methods can be realized with
in-context learning. We choose model-based learning because planned trajectories
make the underlying logic of value-estimates explicit to our foundation model
backbone, providing a concrete instantiation of a trajectory that realizes the
values. This ties into recent work
\parencite{wei_chain_2022,nye_show_2021}
demonstrating that ``chains of thought'' can significantly improve few-shot
performance of foundation models.

Model-based RL requires two ingredients, a rollout-policy used to act during
planning and a world-model used to predict future rewards, terminations, and
states. Since our approach avoids any mutation of the foundation model's
parameters (this would require gradients), we must instead induce the
rollout-policy and the world-model using in-context learning, i.e. by selecting
appropriate prompts. We induce the rollout-policy by prompting the foundation
model with trajectories drawn from the current (or recent) behavior policy
(distinct from the rollout-policy). Similarly, we induce the world-model by
prompting the foundation models with transitions drawn from the agent's history
of experience. Note that our approach assumes access to some translation between
the state-space of the environment and the medium (language, images, etc.) of
the foundation models. This explains how an algorithm might plan and estimate
values using a foundation model. It also explains how the rollout-policy
approximately tracks the behavior policy.

How does the policy improve? When acting in the environment (as opposed to
planning), we choose the action that maximizes the estimated Q-value from the
current state (see \nameref{algo:train} pseudocode, line 6). At time step $t$,
the agent observes the state of the environment (denoted $s_t$) and executes
action $a_t = \argmax_{\Act \in \Actions} \Q^{\Policy_t}(\Obs_t,\Act)$,
where $\Actions = \{\Actions(1),\cdots,\Actions(n)\}$ denotes the set of $n$
actions available, $\Policy_t$ denotes the policy of the agent at time step $t$,
and $\Q^{\Policy}$ denotes the Q-estimate for policy $\Policy$. Taking the
greedy ($\argmax$) actions with respect to $Q^{\pi_t}$ implements a new and
improved policy.

\paragraph{Computing Q-values}\label{para:q-values} This section provides
details on the prompts that we use in our computation of Q-values (see
\nameref{algo:q} pseudocode \& Figure~\ref{fig:q rollouts}). During training, We
maintain a buffer $\Buffer$ of transitions experienced by the agent. To compute
$\Q^{\Policy_t}(\Obs_t, \Act)$ at time step $t$ in the real-world we rollout a
simulated trajectory $\Obs^{1}=\Obs_t$, $\Act^{1} = \Act$, $\Rew^{1}$,
$\Obs^{2}$, $\Act^{2}$, $\Rew^{2}$, $\cdots$, $\Obs^{T}$, $\Act^{T}$,
$\Rew^{T}$, $\Obs^{T+1}$ by predicting, at each simulation time step $u$: reward
$\Rew^{u} \sim \LLM\left( \Buffer_{\Rew},\Obs^{u},\Act^{u} \right)$; termination
$\Ter^{u} \sim \LLM\left( \Buffer_{\Ter},\Obs^{u},\Act^{u} \right)$; observation
$\Obs^{u+1} \sim \LLM\left( \Buffer_{\Obs},\Obs^{u},\Act^{u} \right)$; action
$\Act^{1} \sim \LLM\left( \BestTrajectories,\Obs^{u} \right)$. Termination
$\Ter^{u}$ decides whether the simulated trajectory ends at step $u$.

The prompts $\Buffer_{\Rew}$, $\Buffer_{\Ter}$ contain data sampled from the
replay buffer. For each prompt, we choose some subset of replay buffer
transitions, shuffle them, convert them to text (examples are provided in table
\ref{sec:domains-and-prompt-format}) and clip the prompt at the 4000-token
Codex context limit. We use the same method for $\Buffer_{\Act}$, except that we
use random trajectory subsequences.

In order to maximize the relevance of the prompt contents to the current
inference we select transitions using the following criteria. $\Buffer_{\Ter}$
contains $(\Obs_k, \Act_{k}, \Ter_{k})$ tuples such that $\Act_{k}$ equals
$\Act^{u}$, the action for which the LLM must infer termination.
$\Buffer_{\Rew}$ contains $(\Obs_k, \Act_{k}, \Rew_{k})$ tuples, again
constraining  $\Act_{k} = \Act^{u}$ but also constraining $\Ter_{k} = \Ter^{k}$
--- that the tuple corresponds to a terminal time-step if the LLM inferred
$\Ter^{u} =$ true, and to a non-terminal time-step if $\Ter^{u} =$ false. For
$\Buffer_{\Obs}$, the prompt includes $(\Obs_k, \Act_{k} \Obs_{k+1})$ tuples
with $\Act_{k} = \Act^{u}$ and $\Ter_{k} =$ false (only non-terminal states need
to be modelled).

We also maintain a balance of certain kinds of transitions in the prompt. For
termination prediction, we balance terminal and non-terminal time-steps.
Since non-terminal time-steps far outnumber terminal time-steps, this
eliminates a situation wherein the randomly sampled prompt time-steps are
entirely non-terminal, all but ensuring that the LLM will predict
non-termination. Similarly, for
reward prediction, we balance the number of time-steps corresponding to each
reward value stored in $\Buffer$. In order to balance two collections of
unequal size, we take the smaller and duplicate randomly chosen members
until the sizes are equal.

In contrast to the other predictions, we condition the rollout policy on
trajectory subsequences, not individual time-steps. Prompting with sequences
better enables the foundation model to apprehend the logic behind a policy. Trajectory
subsequences consist of $(\Obs_{k}, \Act_{k})$ pairs, randomly clipped from
the $\RecencyCutoff$ most recent trajectories. More recent trajectories will, in
general demonstrate higher performance, since they come from policies that have
benefited from more rounds of improvement.

In contrast to the other predictions, we condition the rollout policy on
trajectory subsequences, not individual time-steps. Prompting with sequences
better enables the foundation model to apprehend the logic behind a policy. Trajectory
subsequences consist of $(\Obs_{k}, \Act_{k})$ pairs, randomly clipped from
the $\RecencyCutoff$ most recent trajectories. More recent trajectories will, in
general demonstrate higher performance, since they come from policies that have
benefited from more rounds of improvement.

Finally, the Q-value estimate is simply the discounted sum of rewards for the
simulated episode. Given this description of Q-value estimation, we now return
to the concept of policy improvement.

\lstdefinestyle{mystyle}{ basicstyle=\ttfamily\scriptsize }
\lstset{style=mystyle}
\def\graycolor{\color{lightgray}}
\def\blackcolor{\color{black}}

\paragraph{Policy-Improvement} The $\argmax$ (line 6 of
Algorithm~\ref{algo:train}) drives policy improvement in \Algorithm{}.
Critically this is not simply a one-step improvement as with Trajectory
Transformer \parencite{janner_offline_2021} but a mechanism that builds improvement
on top of improvement. This occurs through a cycle in which the $\argmax$
improves behavior. The improved behavior is stored in the buffer $\Buffer$, and
then used to condition the rollout policy. This improves the returns generated
by the LLM during planning rollouts. These improved rollouts improve the
Q-estimates for each action. Completing the cycle, this improves the actions
chosen by the $\argmax$. Because this process feeds into itself, it can drive
improvement without bound until optimality is achieved.

Note that this process takes advantage of properties specific to in-context
learning. In particular, it relies on the assumption that the rollout policy,
when prompted with trajectories drawn from a mixture of policies, will
approximate something like an average of these policies. Given this
assumption, the rollout policy will improve with the improvement of the
mixture of policies from which its prompt-trajectories are drawn. This
results in a kind of rapid policy improvement that works without any use of
gradients.

\paragraph{Prompt-Format} \label{para:prompt-format} The LLM cannot take
non-linguistic prompts, so our algorithm assumes access to a textual
representation of the environment---of states, actions, terminations, and
rewards---and some way to recover the original action, termination, and reward
values from their textual representation (we do not attempt to recover states).
Since our primary results use the Codex language model (see
Table~\ref{tab:llms}), we use Python code to represent these values
(examples are available in Table~\ref{app-tab:promptformat} in the appendix).

In our experiments, we discovered that the LLM world-model was unable to
reliably predict rewards, terminations, and next-states on some of the more
difficult environments. We experimented with providing domain \emph{hints} in
the form of prompt formats that make explicit useful information --- similar to
Chain of Thought Prompting \parencite{wei_chain_2022}. For example, for the
\emph{chain} domain, the hint includes an explicit comparison (\texttt{==} or
\texttt{!=}) of the current state with the goal state. Note that while hints are
provided in the initial context, the LLM must infer the hint content in rollouts
generated from this context.

We use a consistent idiom for rewards and terminations, namely \texttt{assert
  reward == x} and \texttt{assert done} or \texttt{assert not done}. Some
decisions had to be made when representing states and actions. In general, we
strove to use simple, idiomatic, concise Python. On the more challenging
environments, we did search over several options for the choice of hint. We
anticipate that in the future, stronger foundation models will be
increasingly robust to these decisions.

\begin{figure*}[t]
  \centering
  \includegraphics[width=\textwidth]{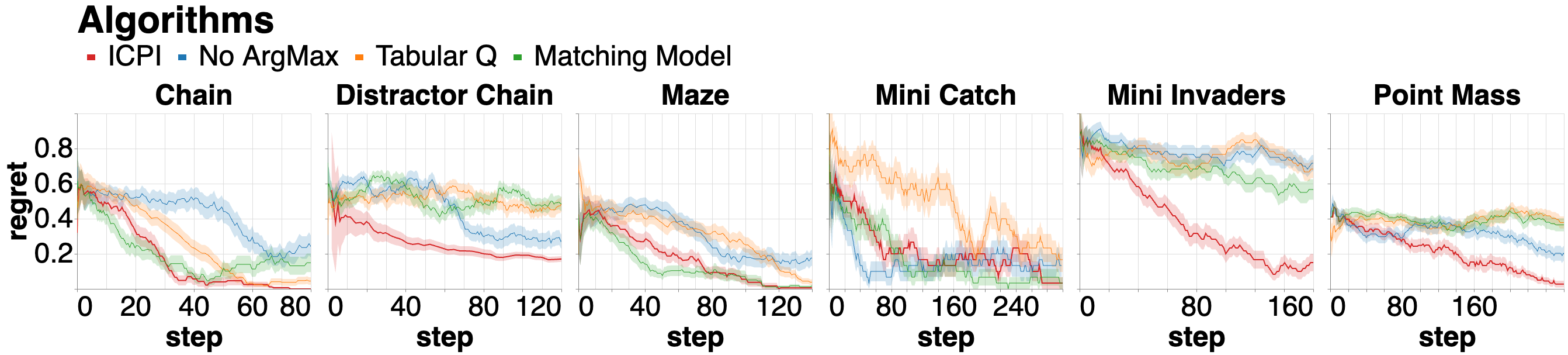}
  %\caption{\textcolor{red}{EVALUATION REGRET}
  \caption{Comparison of \Algorithm{} with three baselines, \NoArgMax{},
    ``Tabular Q,'' and ``Nearest Neighbor.'' The $y$-axis depicts regret (normalized between 0 and 1),
    computed relative to an optimal return with a discount-factor of 0.8.
    The $x$-axis depicts time-steps during training. Error bars
    are standard errors from four seeds.}
  \label{fig:algorithms}
\end{figure*}

\section{Experiments} %\todo{"Results" or "Experiments"?}
\label{sec:experiments}

We have three main goals in our experiments: (1) Demonstrate that the agent
algorithm can in fact quickly learn good policies,  using pretrained LLMs, in a
set of six simple illustrative domains of increasing challenge; (2) provide
evidence through an ablation that the policy-improvement step---taking the
$\argmax$ over Q-values computed through LLM rollouts---accelerates learning;
and (3) investigate the impact of using different LLMs (see
Table~\ref{tab:llms})---different sizes and trained on different data, in
particular, trained on (mostly) natural language (GPT-3 and GPT-J) vs.\ program
code (Codex and InCoder). We next describe the six domains and their
associated prompt formats, and then describe the experimental methodology and
results.

\subsection{Domains and prompt format}
\label{sec:domains-and-prompt-format}

\paragraph{Chain.}~~ In this environment, the agent occupies an 8-state chain.
The agent has three actions:  \emph{Left}, \emph{right}, and \emph{try goal}.
%\todo{Should these be texttt format or are we intentionally trying to
%distinguish between the representation and the underlying action?} \todo{I was
%trying to distinguish between real-world actions and text encoding, we reserve
%texttt for latter. -rl}
The \emph{try goal} action always terminates the episode, conferring a reward of
1 on state 4 (the goal state) and 0 on all other states.
%Because this environment has simpler transitions than the other two, we see the
%clearest evidence of learning here. Note that the initial batch of successful
%trajectories collected from random behavior will usually be suboptimal, moving
%inefficiently toward the goal state. We include a discount value of 0.8 in our
%diagram to show the improvement in efficiency of the policy learned by the
%agent over the course of training. \emph{Prompt format.}
Episodes also terminate after 8 time-steps. States are represented as numbers
from 0 to 7, as in \texttt{assert state == n}, with the appropriate integer
substituted for \texttt{n}. The actions are represented as functions
\texttt{left()}, \texttt{right()}, and \texttt{try\_goal()}. For the hint, we
simply indicate whether or not the current state matches the goal state, 4.
% As our results indicate (figure~\ref{fig:ablations}), this actually had
% little effect on performance.

\paragraph{Distractor Chain.} This environment is an 8-state chain, identical to
the \emph{chain} environment, except that the observation is a \emph{pair} of
integers, the first indicating the true state of the agent and the second acting
as a distractor which transitions randomly within $\{0, \dots, 7\}$. The agent
must therefore learn to ignore the distractor integer and base its inferrences
on the information contained in the first integer. Aside from the addition of
this distractor integer to the observation, all text representations and hints
are identical to the \emph{chain} environment.

\paragraph{Maze.} The agent navigates a small $3\times 3$ gridworld with
obstacles. The agent can move \emph{up}, \emph{down}, \emph{left}, or
\emph{right}. The episode terminates with a reward of 1 once the agent navigates
to the goal grid, or with a reward of 0 after 8 time-steps. This environment
tests our algorithms capacity to handle 2-dimensional movement and obstacles, as
well as a 4-action state-space. We represent the states as namedtuples ---
\texttt{C(x, y)}, with integers substituted for \texttt{x} and \texttt{y}.
Similar to \emph{chain}, the hint indicates whether or not the state corresponds
to the goal state.
% , and again, we found in our results that this hinting
% mechanism had little effect.

\paragraph{Mini Catch.}
%In this environment,
The agent operates a paddle to catch a falling ball. The ball falls from a
height of 5 units, descending one unit per time step. The paddle can \emph{stay}
in place (not move), or move \emph{left} or \emph{right} along the bottom of the
4-unit wide plane. The agent receives a reward of 1 for catching the ball and 0
for other time-steps. The episode ends when the ball's height reaches the paddle
regardless of whether or not the paddle catches the ball.
We chose this environment specifically to challenge the
action-inference/rollout-policy component of our algorithm. Specifically, note
that the success condition in Mini Catch allows the paddle to meander before
moving under the ball---as long as it gets there on the final time-step.
Successful trajectories that include movement \textit{away} from the ball thus
make a good rollout policies more challenging to learn (i.e., elicit from the
LLM via prompts).

Again, we represent both the paddle and the ball as namedtuples \texttt{C(x,
  y)} and we represent actions as methods of the \texttt{paddle} object:
\texttt{paddle.stay()}, \texttt{paddle.left()}, and \texttt{paddle.right()}.
For the hint, we call out the location of the paddle's $x$-position, the
ball's $x$-position, the relation between these positions (which is larger
than which, or whether they are equal) and the ball's $y$-position.
Table~\ref{app-tab:promptformat} in the appendix provides an example. We also include the text
\texttt{ball.descend()} to account for the change in the ball's position
between states. \paragraph{Mini Invaders.}~~ The agent operates a ship that
shoots down aliens which descend from the top of the screen. At the beginning
of an episode, two aliens spawn at a random location in two of four columns.
The episode terminates when  an alien reaches the ground (resulting in 0
reward) or when the ship shoots down both aliens (the agent receives 1 reward
per alien). The agent can move \emph{left}, \emph{right}, or \emph{shoot}.
This domain highlights \Algorithm{}'s capacity to learn incrementally, rather
than discovering an optimal policy through random exploration and then
imitating that policy, which is how our \NoArgMax{} baseline learns (see
\nameref{para:baselines}). \Algorithm{} initially learns to shoot down one
alien, and then builds on this good but suboptimal policy to discover the
better policy of shooting down both aliens. In contrast, random exploration
takes much longer to discover the optimal policy and the \NoArgMax{} baseline
has only experienced one or two successful trajectories by the end of
training.

We represent the ship by its namedtuple coordinate (\texttt{C(x, y)}) and the
aliens as a list of these namedtuples. When an alien is shot down, we substitute
\texttt{None} for the tuple, as in \texttt{aliens == [C(x, y), None]}. We add
the text \texttt{for a in aliens: a.descend()} in order to account for the
change in the alien's position between states.

\paragraph{Point-Mass.}~~ A point-mass spawns at a random position on a
continuous line between $-6$ and $+6$ with a velocity of 0. The agent can either
\emph{accelerate} the point-mass (increase velocity by 1) or \emph{decelerate}
it (decrease the velocity by 1). The point-mass position changes by the amount
of its velocity each timestep. The episode terminates with a reward of 1
once the point-mass is between $-2$ and $+2$ and its velocity is 0 once again.
The episode also terminates after 8 time-steps. This domain tests the
algorithm's ability to handle continuous states.

States are represented as \texttt{assert pos == p and vel == v}, substituting
floats rounded to two decimals for \texttt{p} and \texttt{v}. The actions
are \texttt{accel(pos, vel)} and \texttt{decel(pos, vel)}. The hint
indicates whether the success conditions are met, namely the relationship
of \texttt{pos} to $-2$ and $+2$ and the whether or not \texttt{vel == 0}.
The hint includes identification of the aliens' and the ship's $x$-positions
as well as a comparison between them.

\subsection{Experiment Methodology and Results}

\paragraph{Methodology and evaluation.}
\label{para:methodology}
For the results, we record the agent's regret over the course of training
relative to an optimal policy computed with a discount factor of 0.8. For all
experiments $\RecencyCutoff = 8$ (the number of most recent successful
trajectories to include in the prompt). We did not have time for hyperparameter
search and chose this number based on intuition, however the $\RecencyCutoff =
  16$ baseline demonstrates results when this hyperparameter is doubled. All
results use 4 seeds.

For both versions of Codex, we used the OpenAI Beta under the API Terms of Use.
For GPT-J~\parencite{wang_gpt-j-6b_2021} , InCoder~\parencite{fried_incoder_2022} and
OPT-30B~\parencite{zhang_opt_2022}, we used the open-source implementations from
Huggingface Transformers \parencite{wolf_transformers_2020}, each running on one
Nvidia A40 GPU. All language models use a sampling temperature of 0.1.

\paragraph{Comparison of \Algorithm{} with baseline
  algorithms.}\label{para:baselines} We compare \Algorithm{} with three
baselines (Fig. \ref{fig:algorithms}).

The \NoArgMax{} baseline learns a good policy through random exploration and
then imitates this policy. This baseline assumes access to a ``success threshold''
for each domain --- an undiscounted cumulative return greater than which a
trajectory is considered successful. The action selection mechanism emulates
ICPI's rollout policy: prompting the LLM with a set of trajectories and
eliciting an action as output. For this baseline, we only include trajectories
in the prompt whose cumulative return exceeds the success threshold. Thus the
policy improves as the number of successful trajectories in the prompt increases
over time. Note that at the start of learning, the agent will have experienced
too few successful trajectories to effectively populate the policy prompt. In
order to facilitate exploration, we act randomly until the agent experiences 3
successes.

``Tabular Q'' is a standard tabular Q-learning algorithm, which uses a learning rate of $1.0$ and optimistically initializes the Q-values to $1.0$.

\MatchingModel{} is a baseline which uses the trajectory history instead of an
LLM to perform modelling. This baseline searches the trajectory buffer for the
most recent instance of the current state, and in the case of
transition/reward/termination prediction, the current action. If a match is
found, the model's outputs the historical value (e.g. the reward associated with
the state-action pair found in the buffer). If no match is found, the modelling
rollout is terminated. Recall that ICPI breaks ties randomly during action
selection so this will often lead to random action selection.

As our results demonstrate, only \Algorithm{} learns good policies on all
domains. We attribute this advantage to \Algorithm{}'s ability to generalize
from its context to unseen states and state/action pairs (unlike \TabularQ{} and
\MatchingModel{}). Unlike \NoArgMax{} \Algorithm{} is able to learn progressively,
improving the policy before experiencing good trajectories.

\begin{figure*}[t]
  \centering
  \includegraphics[width=\textwidth]{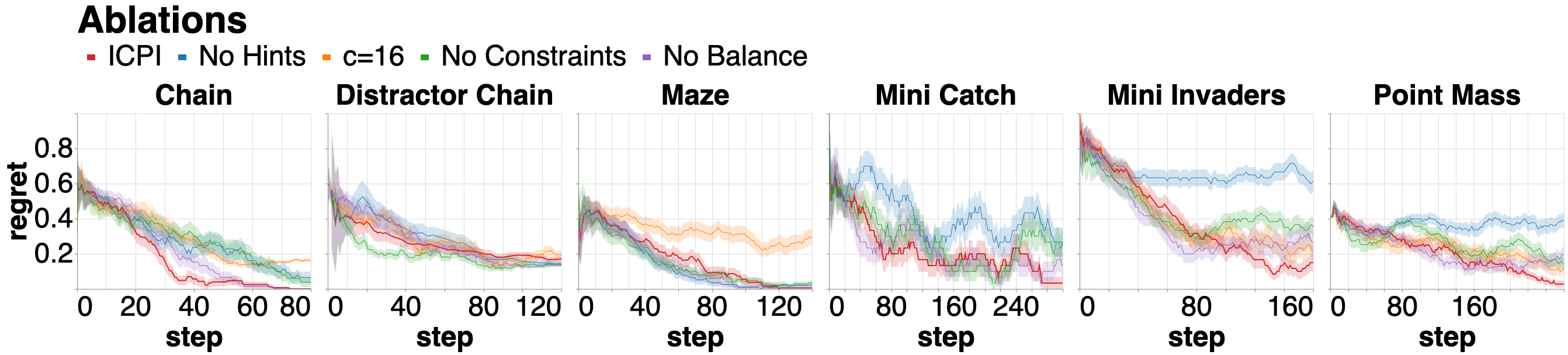}
  %\caption{\textcolor{red}{EVALUATION REGRET}
  \caption{Comparison of \Algorithm{} with ablations. The $y$-axis depicts
    regret (normalized between 0 and 1), computed relative to an optimal
    return with a discount-factor of 0.8. The $x$-axis depicts time-steps
    during training. Error bars are standard errors from four seeds.}
  \label{fig:ablations}
\end{figure*}
\paragraph{Ablation of \Algorithm{} components.} With these experiments, we
ablate those components of the algorithm which are not, in principle, essential
to learning (Fig. \ref{fig:ablations}). \NoHints{} ablates the hints described in
the \nameref{para:prompt-format} paragraph. \NoBalance{} removes the balancing
of different kinds of time-steps described in the \nameref{para:q-values}
paragraph (for example, $\Buffer_{\Ter}$ is allowed to contain an unequal number
of terminal and non-terminal time-steps). The \NoConstraints{} baseline removes
the constraints on these time-steps described in the same paragraph. For
example,
% $\Buffer_\Done$ is allowed to contain time-steps with actions other
% than $\ActionI[u]$ (the action for which the inference is being performed) and
$\Buffer_{\Rew}$ is allowed to contain a mixture of terminal and non-terminal
time-steps (regardless of the model's termination prediction). Finally, \CSixteen{} prompts the rollout
policy with the last 16 trajectories (instead of the last 8, as in
\Algorithm{}). We find that while some ablations match \Algorithm{}'s
performance in several domains, none match its performance on all six.
% The
% \NoBalance{} and \NoConstraints{} baselines still appear to be capable of
% learning, but they are generally outperformed by \Algorithm{}, especially on
% the more difficult domains. Ablating constraints appears to be more
% deleterious than ablating balance.  

\begin{figure*}
  \centering
  \includegraphics[width=\textwidth]{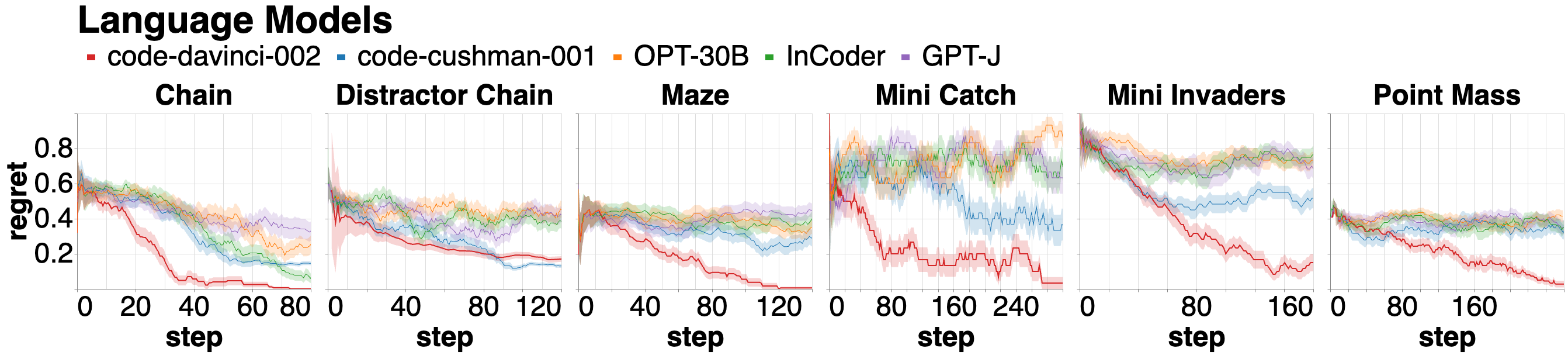}
  %\caption{\textcolor{red}{EVALUATION REGRET}
  \caption{Comparison of \Algorithm{} with ablations. The $y$-axis depicts
    regret (normalized between 0 and 1), computed relative to an optimal
    return with a discount-factor of 0.8. The $x$-axis depicts time-steps of
    training. Error bars are standard errors from four seeds.}
  \label{fig:language-models}
\end{figure*}

\paragraph{Comparison of Different Language Models.} While our lab lacks the
resources to do a full study of scaling properties, we did compare several
language models of varying size (Fig. \ref{fig:language-models}). See Table \ref{tab:llms} for details about
these models. Both \texttt{code-davinci-002} and \texttt{code-cushman-001} are
variations of the Codex language model. The exact number of parameters in these
models is proprietary according to OpenAI, but \cite{chen_evaluating_2021}
describes Codex as fine-tuned from GPT-3 \cite{brown_language_2020}, which
contains 185 billion parameters. As for the distinction between the variations,
the OpenAI website describes \texttt{code-cushman-001} as ``almost as capable as
Davinci Codex, but slightly faster.''

We found that none of the smaller models were capable of
demonstrating learning on any domain but the simplest, \emph{chain}.
Examining the trajectories generated by agents trained using these models, we
noted that in several cases, they seemed to struggle to apprehend the
underlying ``logic'' of successful trajectories, which hampered the ability
of the rollout policy to produce good actions.
% We also noticed that hints
% failed to have the same benefit for these smaller models, suggesting that
% they were unable to perform ``Chain-of-Thought'' reasoning as effectively as
% the larger model.
Since these smaller models were not trained on identical data, we are unable to
isolate the role of size in these results. However, the failure of all of these
smaller models to learn suggests that size has some role to play in performance.
We conjecture that larger models developed in the future may demonstrate
comparable improvements in performance over our Codex model.

\paragraph{Limitations} In principle, ICPI can work on any control task with a
discrete state space. All that the method requires is a sequence prediction
model (in the paper we use Codex) capable of inferring state transitions and
action probabilities given a history of behavior. Such a model will ensure that
the rollouts (described in Computing Q-values) are unbiased monte-carlo
estimates of value for the current policy. Given such estimates, the algorithm
will implement policy iteration, a method which is proven to converge. The
domains in our paper were limited by the abilities of Codex to accurately
predict transitions and actions in more complex domains. As sequence models
mature, we anticipate that more domains will become tractable for ICPI.

\paragraph{Societal Impacts}\label{para:impacts} An extensive literature has
explored the possible positive and negative impacts of LLMs. Some of this work
has explored mitigation strategies. In extending LLMs to RL, our work inherits
these benefits and challenges. We highlight two concerns: the use of LLMs to
spread misinformation and the detrimental carbon cost of training and using
these models.

\begin{table*}[t]
  \centering
  \caption{Pretrained Large Language Models (LLMs) Used in Experiments}
  \label{tab:llms}

  \begin{tabular}{m{2in}cm{2.6in}}
    \toprule

    \textbf{Model}                           & \textbf{Parameters} & \textbf{Training data}                                   \\

    \midrule
    {GPT-J} \parencite{wang_gpt-j-6b_2021}   & 6 billion           & ``The Pile'' \parencite{gao_pile_2020}, an 825GB English
    corpus incl.\ Wikipedia, GitHub, academic pubs                                                                            \\

    \hdashline

    {InCoder} \parencite{fried_incoder_2022} & 6.7 billion         & 159 GB of open-source StackOverflow code                 \\
    \hdashline
    {OPT-30B} \parencite{zhang_opt_2022}     & 30 billion          & 180B tokens of predominantly English data
    including ``The Pile'' \parencite{gao_pile_2020} and ``PushShift.io Reddit'' \parencite{baumgartner_pushshift_2020}
    \\
    \hdashline

    {Codex} \parencite{chen_evaluating_2021} & 185 billion         & 179 GB of GitHub code                                    \\
    \bottomrule
  \end{tabular}
\end{table*}

\section{Conclusion}

Our main contribution is a method for implementing policy iteration algorithm
using Large Language Models and the mechanism of in-context learning. The
algorithm uses a foundation models as both a world model and policy to compute
Q-values via rollouts. Although we presented the method here as text-based, it
is general enough to be applied to any foundation models that works through
prompting, including multi-modal models like \cite{reed_generalist_2022} and
\cite{seo_harp_2022}. In experiments we showed that the algorithm works in six
illustrative domains imposing different challenges for ICPI, confirming the
benefit of the LLM-rollout-based policy improvement. While the empirical results
are quite modest, we believe the approach provides an important new way to use
LLMs that will increase in effectiveness as the models themselves become better,
as our size comparison study suggests.

\newpage
\section{NeurIPS Paper Checklist}
\subsection{For all authors\dots}
\begin{enumerate}
  \item \textbf{Do the main claims made in the abstract and introduction accurately reflect the paper's contributions and scope?}
        Yes. Our claim about a novel algorithm is substantiated in the \hyperref[sec:related]{Related Works} and \hyperref[sec:algorithms]{Algorithms} section. Our claim that it can produce improved behavior is substantiated in the \hyperref[sec:experiments]{Experiments} section.
  \item \textbf{Did you describe the limitations of your work?}
        Yes. We have a \hyperref[para:limitations]{Limitations} section that addresses this directly.
  \item \textbf{Did you discuss any potential negative societal impacts of your work?}
        Yes. We have a \hyperref[para:impacts]{Societal Impacts} section that addresses this directly.
  \item \textbf{Have you read the ethics review guidelines and ensured that your paper conforms to them?} Yes.
\end{enumerate}
\subsection{If you are including theoretical results\dots}
This paper does not include theoretical results.
\subsection{If you ran experiments\dots}
\begin{enumerate}
  \item \textbf{Did you include the code, data, and instructions needed to reproduce the main experimental results (either in the supplemental material or as a URL)?}
        Yes. This information is included in the supplemental material
  \item \textbf{Did you specify all the training details (e.g., data splits, hyperparameters, how they were chosen)?}
        Yes. This information is described in the \hyperref[para:methodology]{Methodology and Evaluation} section of our paper.
  \item \textbf{Did you report error bars (e.g., with respect to the random seed after running experiments multiple times)?}
        Yes. We report error bands for all graphs.
  \item \textbf{Did you include the amount of compute and the type of resources used (e.g., type of GPUs, internal cluster, or cloud provider)?}
        Yes. This information is present in the \hyperref[para:methodology]{Methodology and Evaluation} section of our paper. We were not able to compute an estimate of our CO2 emissions, since we do not have access to the details of the OpenAI machines that were accessed through the API.
\end{enumerate}
\subsection{If you are using existing assets (e.g., code, data, models) or curating/releasing new assets\dots}
\begin{enumerate}
  \item \textbf{If your work uses existing assets, did you cite the creators?} Yes. In our   \hyperref[tab:llms]{Pretrained Large Language Models (LLMs) Used in Experiments} table, we cited all of the authors of the original models as well as the authors of the datasets that those models used.
  \item \textbf{Did you mention the license of the assets?} Yes. The only asset that we are using that is subject to a license is the OpenAI API. We link to the Terms of Use \hyperref[para:methodology]{Methodology and Evaluation} section.
  \item \textbf{Did you include any new assets either in the supplemental material or as a URL?} Yes. Our source code is included in the supplemental material.
  \item \textbf{Did you discuss whether and how consent was obtained from people whose data you're using/curating?} No. We are not using any data derived directly from human sources.
  \item \textbf{Did you discuss whether the data you are using/curating contains personally identifiable information or offensive content?} No.  We are not using any data derived directly from human sources.
\end{enumerate}
\subsection{If you used crowdsourcing or conducted research with human subjects\dots}
This paper does not use crowdsourcing or conduct research with human subjects.

\printbibliography{}

\end{document}